\documentclass{article}

\usepackage{arxiv}

\usepackage[utf8]{inputenc} 
\usepackage[T1]{fontenc}    
\usepackage{hyperref}       
\usepackage{url}            
\usepackage{booktabs}       
\usepackage{amsfonts}       
\usepackage{nicefrac}       
\usepackage{microtype}      
\usepackage{lipsum}
\usepackage{graphicx}
\usepackage{cite}
\usepackage{amsmath,amssymb,amsfonts}
\usepackage{algorithmic}
\usepackage{graphicx}
\usepackage{textcomp}
\usepackage{multirow}
\usepackage{amsmath}
\usepackage{amsfonts}
\usepackage{enumitem}
\usepackage{balance}
\usepackage{csquotes}
\usepackage{subcaption}
\usepackage{cleveref}
\usepackage{hyperref}
\usepackage{float}
\graphicspath{ {./images/} }

\begin{document}

\thispagestyle{empty}
{\Huge \textbf{IEEE Copyright Notice}}\\[1em]
{\large
\noindent
\textcopyright~2024 IEEE. Personal use of this material is permitted. Permission from IEEE must be obtained for all other uses, in any current or future media, including reprinting/republishing this material for advertising or promotional purposes, creating new collective works, for resale or redistribution to servers or lists, or reuse of any copyrighted component of this work in other works.

DOI: \href{https://doi.org/10.1109/ICCIT64611.2024.11022039}{10.1109/ICCIT64611.2024.11022039}
}
\newpage

\fancypagestyle{titlepage}{
  \fancyhf{}
  \fancyhead[C]{\footnotesize This work has been accepted for publication in 2024 27th International Conference on Computer and Information Technology (ICCIT).\\
  The final published version will be available via IEEE Xplore.\\
  DOI: \href{https://doi.org/10.1109/ICCIT64611.2024.11022039}{10.1109/ICCIT64611.2024.11022039}
  }
  \renewcommand{\headrulewidth}{0pt}
}

\title{An Exploratory Approach Towards Investigating and Explaining Vision Transformer and Transfer Learning for Brain Disease Detection}

\author{
 Shuvashis Sarker \\
  Department of Computer Science and Engineering\\
  Ahsanullah University of Science and Technology\\
  Dhaka,Bangladesh\\
  \texttt{shuvashisofficial@gmail.com} \\
   \And
Shamim Rahim Refat \\
  Department of Computer Science and Engineering\\
  Ahsanullah University of Science and Technology\\
  Dhaka,Bangladesh\\
  \texttt{n.a.refat2000@gmail.com} \\
  \And
 Faika Fairuj Preotee \\
  Department of Computer Science and Engineering\\
  Ahsanullah University of Science and Technology\\
  Dhaka,Bangladesh\\
  \texttt{faikafairuj2001@gmail.com} \\
 \And
 Shifat Islam \\
  Department of Computer Science and Engineering\\
  Bangladesh University of Engineering and Technology\\
  Dhaka,Bangladesh\\
  \texttt{shifat.islam.buet@gmail.com} \\
  \And
Tashreef Muhammad \\
  Department of Computer Science and Engineering\\
  Southeast University\\
  Dhaka,Bangladesh\\
  \texttt{tashreef.muhammad@gmail.com} \\
  \And
 Mohammad Ashraful Hoque \\
  Department of Computer Science and Engineering\\
 Southeast University\\
  Dhaka,Bangladesh\\
  \texttt{ashraful@seu.edu.bd} \\
}

\maketitle
\thispagestyle{titlepage}

\begin{abstract}
The brain is a highly complex organ that manages many important tasks, including movement, memory and thinking. Brain-related conditions, like tumors and degenerative disorders, can be hard to diagnose and treat. Magnetic Resonance Imaging (MRI) serves as a key tool for identifying these conditions, offering high-resolution images of brain structures. Despite this, interpreting MRI scans can be complicated. This study tackles this challenge by conducting a comparative analysis of Vision Transformer (ViT) and Transfer Learning (TL) models such as VGG16, VGG19, Resnet50V2, MobilenetV2 for classifying brain diseases using MRI data from Bangladesh based dataset. ViT, known for their ability to capture global relationships in images, are particularly effective for medical imaging tasks. Transfer learning helps to mitigate data constraints by fine-tuning pre-trained models. Furthermore, Explainable AI (XAI) methods such as GradCAM, GradCAM++, LayerCAM, ScoreCAM, and Faster-ScoreCAM are employed to interpret model predictions. The results demonstrate that ViT surpasses transfer learning models, achieving a classification accuracy of 94.39\%. The integration of XAI methods enhances model transparency, offering crucial insights to aid medical professionals in diagnosing brain diseases with greater precision.
\end{abstract}

\keywords{Vision Transformer (ViT) \and Transfer Learning (TL) \and Explainable AI (XAI) \and Magnetic Resonance Imaging (MRI) \and State-of-the-Art (SOTA)}

\section{Introduction}
The human brain serves as the central nervous system, directing and coordinating all physiological processes and functions. Brain diseases, including tumors and neurodegenerative conditions, pose substantial hurdles in terms of both diagnosis and treatment. Magnetic Resonance Imaging (MRI) has emerged as a crucial diagnostic tool for identifying brain disorders, providing highly detailed images of soft tissues and enabling specialists to detect abnormalities with remarkable precision. However, the complex nature of brain structures makes it challenging to interpret MRI data accurately.

In recent years, advancements in Artificial Intelligence (AI), particularly deep learning techniques, have revolutionized medical imaging. Vision Transformers (ViT) outperform in medical image analysis by leveraging self-attention processes to capture global dependencies, rendering them suitable for applications like as brain tumor classification. However, ViT requires extensive datasets, posing a challenge in areas with constrained medical imaging data, such as Bangladesh\cite{jaffar2024combining}. Transfer learning (TL) mitigates this issue by refining pretrained models on limited medical datasets, thereby strengthening brain disease identification and decreasing the necessity for considerable expert involvement, which ultimately improves clinical decision-making.

This study employs Vision Transformers (ViT) to diagnose brain disorders using MRI data from a Bangladeshi population. Transfer learning is utilized to address data shortages by refining pre-trained models, hence enhancing performance with less training data. Explainable AI (XAI) techniques are employed to identify critical MRI regions that affect predictions, hence improving model transparency. The research compares the efficacy of ViT and transfer learning models to determine their robustness and generalizability in medical imaging.
The primary objectives of this study are:

\begin{enumerate}[label=\roman*.] \item To develop a brain MRI classification framework using a localized Bangladeshi dataset collected from the National Institute of Neurosciences \& Hospital (NINS). \item To improve classification accuracy by incorporating the Vision Transformer (ViT) model, which has not been previously applied to this dataset, leveraging its self-attention processes for enhanced feature extraction.\item To compare the performance of ViT with other transfer learning models to ensure robustness and evaluate their effectiveness on medical imaging. \item To apply multiple Explainable AI (XAI) techniques to enhance the interpretability of model predictions, facilitating clearer insights into the classification process. \end{enumerate}
The remainder of this paper is structured as follows: Section \ref{Related Work} discusses the related work, while Section \ref{Methodology} outlines the proposed methodology, and Section \ref{Result Analysis} presents the analysis of the results. Techniques related to Gradient Based Explainable AI are detailed in Section \ref{Gradient Based Explainable AI}, Limitations and Future Works are described in Section \ref{Limitations and Future Work} and concluding remarks are provided in Section \ref{Conclusion}.

\section{Related Works}
\label{Related Work}
\subsection{Vision Transformer(ViT)}
In recent years, Vision Transformers (ViTs) have gained significant attention for medical image classification tasks. Van Dongen \cite{vancomparison} explored the use of Vision Transformers (ViTs) for brain tumor classification using MRI images. It tested four ViT models (ViT-B/16, ViT-B/32, ViT-L/16, and ViT-L/32) and introduced a "model soup" approach, averaging the weights of multiple fine-tuned models. The best individual model, ViT-L/32, achieved 95.31\% accuracy, while the Combi Soup of ViT-L/16 models outperformed it with 96.12\% accuracy in validation. A new way to classify brain tumors was created by Jaffar\cite{jaffar2024combining} that combines iResNet and Vision Transformers (ViTs). The model used iResNet to get local feature extraction and ViTs to get global environmental information. An attention-based feature merge module optimized the accuracy of classification. The suggested model was 99.2\% accurate, which was much better than other models like InceptionV3, ResNet, and DenseNet. Similarly, Aloraini et al. \cite{aloraini2023combining} introduced a hybrid Transformer-Enhanced Convolutional Neural Network (TECNN) for brain tumor classification using MRI images. The model combines CNN for local feature extraction and transformers for global context. It outperformed several state-of-the-art methods, achieving 96.75\% accuracy on the  \enquote{\textit{BraTS 2018}} dataset and 99.10\% accuracy on the \enquote{\textit{Figshare}} dataset.
Krishnan et al.\cite{krishnan2024enhancing} presented a Rotation Invariant Vision Transformer (RViT) for the classification of brain tumors utilizing MRI data. The model adeptly managed changes in image orientation by including rotational patch embeddings. The RViT surpassed baseline ViTs and other leading methodologies, with an accuracy of 98.6\% with flawless sensitivity and elevated specificity.

\subsection{Transfer Learning with XAI}
Transfer learning utilizing Explainable AI(XAI) integrates the efficacy of pre-trained models and fine-tuning with interpretative strategies. Zeineldin et al.\cite{zeineldin2022explainability} introduced the NeuroXAI framework to enhance the explainability of deep learning models for brain tumor analysis. Using XAI techniques like Grad-CAM and SmoothGrad, they provided visual explanations for model predictions. Their ResNet-50 model achieved 98.62\% accuracy, while the segmentation model demonstrated strong performance with a 92\% Dice score for tumor segmentation. Moreover, Hasan et al.\cite{hasan2023deep} created a brain tumor classification system utilizing transfer learning models on the BrainTumorInSight dataset. The ResNet50 model attained an accuracy of 96.3\%, utilizing LIME to offer visual elucidation of critical MRI regions that affect the model's determinations.
 Another researcher Shad et  al.\cite{shad2021exploring} employed a pipeline of ResNet-50, VGG16, and Inception V3 to categorize phases of Alzheimer’s disease with T1-weighted MRI images from a hybrid Kaggle dataset. The maximum categorical accuracy of 86.82\% was attained with VGG16, with XAI techniques such as LIME applied for interpretability. Furthermore, Mahmud et al.\cite{mahmud2024explainable} developed a model for Alzheimer’s disease diagnosis using the OASIS-2 MRI dataset and transfer learning with VGG16 and DenseNet169. The model achieved 96\% accuracy, leveraging saliency maps and Grad-CAM for enhanced interpretability and clinical application.

\section{Methodology}
\label{Methodology}
\subsection{Dataset}
The Dataset which we have utilized in this study was jointly developed by the National Institute of Neuroscience (NINS), Bangladesh, and the Computer Science and Engineering department of the University of Dhaka\cite{brima2021deep}. This Dataset is publicly available at Figshare\cite{Brima2021BrainMRI} for further study. The dataset consists of 5,285 T1-weighted contrast-enhanced brain MRI images, with various pixel resolution.
\begin{figure}[ht]
\centering
\includegraphics[width=\linewidth]{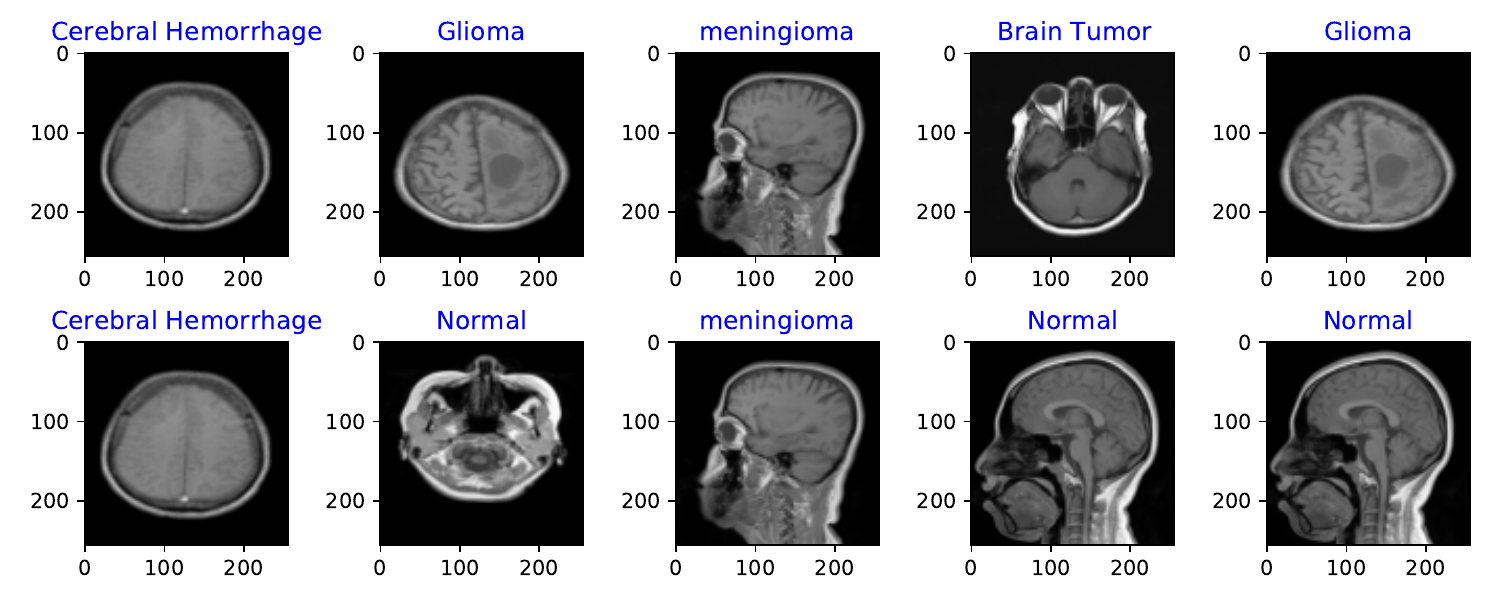}
\caption{Sample Images of the dataset.}
\label{Figure 1}
\end{figure}
This dataset is visually represented in Figure \ref{Figure 1}, which showcases a selection of random images from the dataset. These images are classified into 37 distinct categories. The extensive variety of categories in the dataset facilitates the creation of sophisticated diagnostic algorithms. This invaluable resource plays a crucial role in the training and validation of machine learning models that utilize computational techniques in medical imaging to enhance the precision and effectiveness of neurological diagnostics. 

\begin{figure}[htbp]
    \centering
    \includegraphics[width=\linewidth]{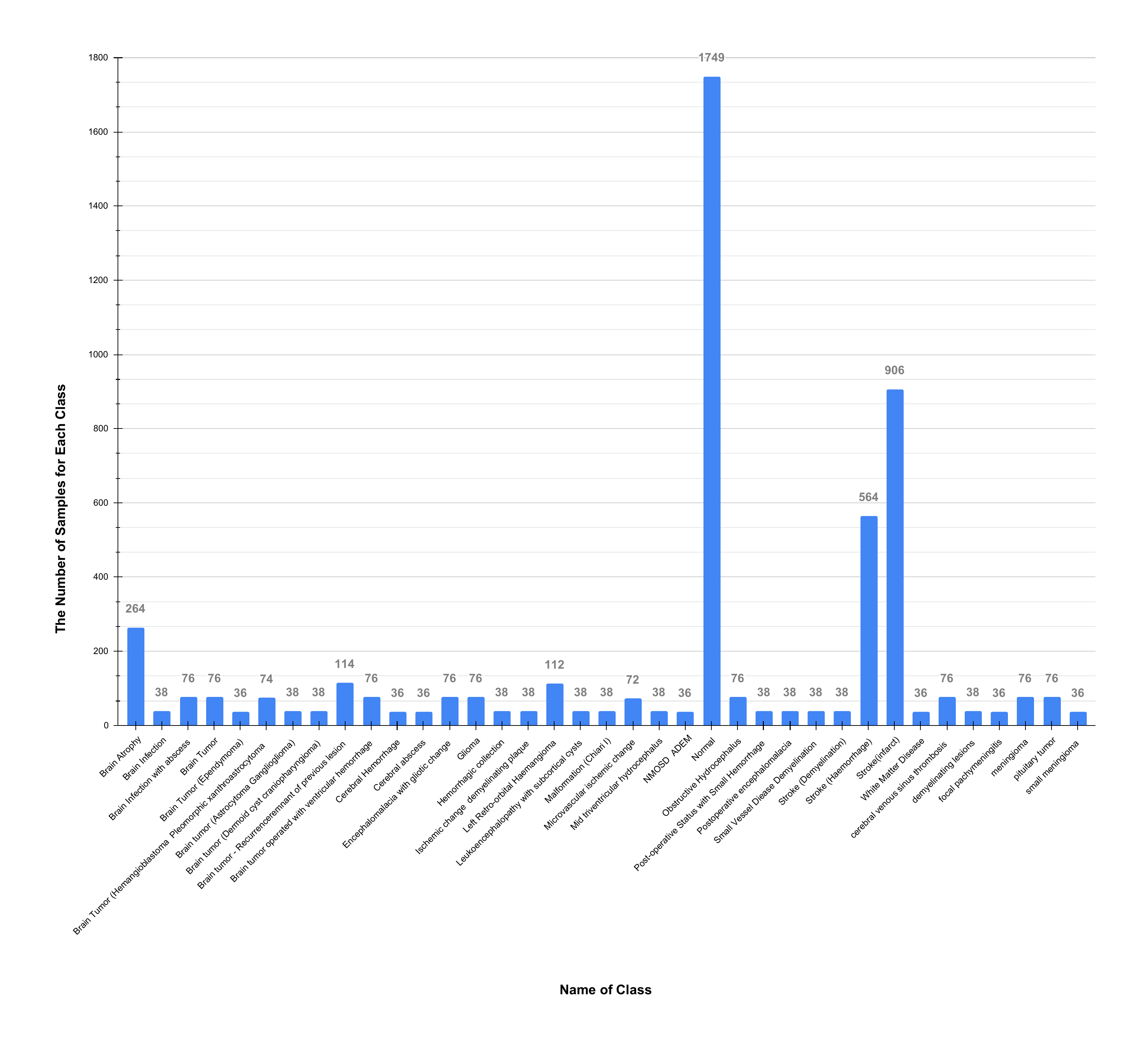}
    \caption{Distribution of Classes in the MRI Dataset}
    \label{Figure 2}
\end{figure}

Figure \ref{Figure 2} depicts a bar plot that visually illustrates the dataset comprising 37 distinct classes, effectively depicting the quantity of MRI pictures for each class.
\subsection{Data Preprocessing}

\begin{figure}[ht]
\centering
\includegraphics[width=\linewidth]{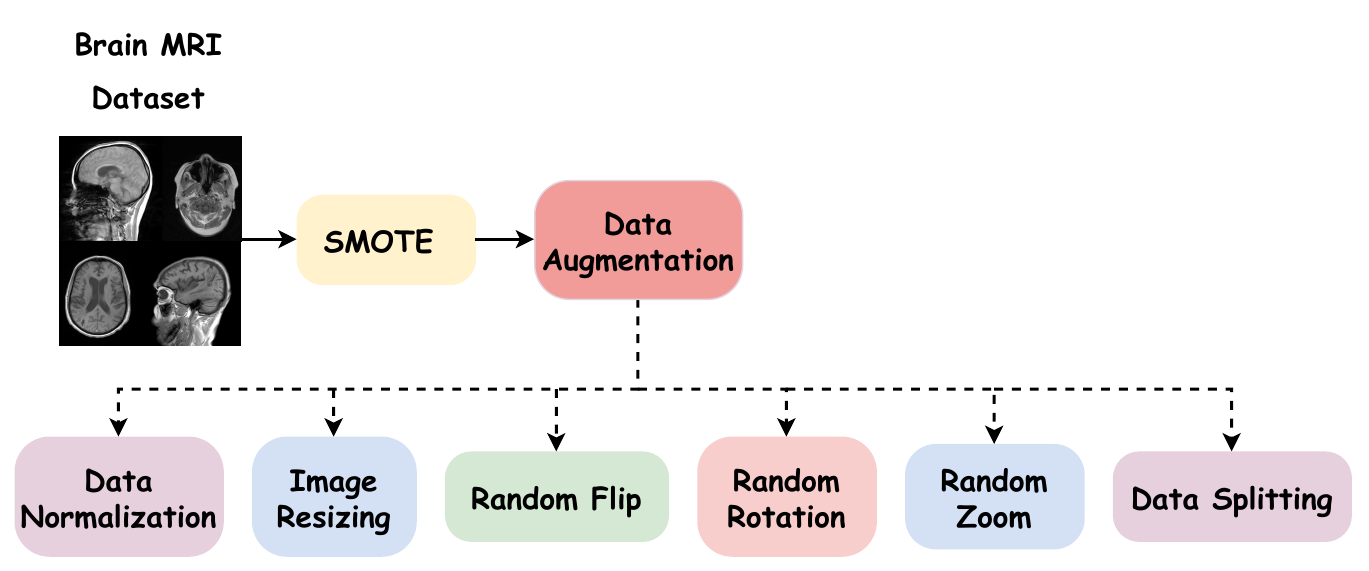}
\caption{Overview of the data preprocessing steps}
\label{Figure 3}
\end{figure}

In the data preprocessing stage, we have at first selected brain MRI images from the dataset to ensure consistency in how the model processes the images during training and testing. To rectify class imbalance, we have employed SMOTE \cite{chawla2002smote}, whereby the minority classes are augmented to align with the mean sample count across all classes, determined by aver aging the complete dataset. \Cref{Figure 3} provides a comprehensive overview of the data preprocessing steps which are involved in this process.

\par Next, we have implemented a data augmentation pipeline
using TensorFlow to introduce variability in the training set.
This process included normalization, resizing the images to
128x128, random horizontal flipping, random rotations with a
reduced factor of 0.01, and random zooming with height and width factors of 0.05. These augmentation techniques helped
the model generalize better by exposing it to diverse versions
of the images. Finally, the dataset was split into training,
validation, and test sets with an 80:10:10 ratio to ensure
balanced and unbiased evaluation of the model’s performance.

\begin{figure*}[ht]
\centering
\includegraphics[width=\linewidth]{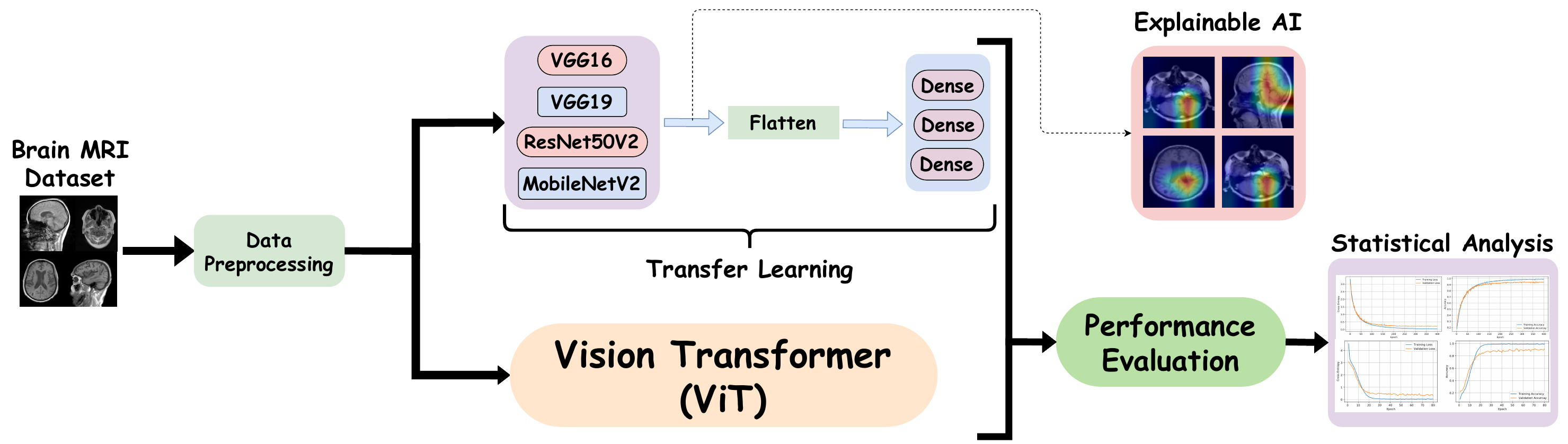}
\caption{Proposed Model Architecture}
\label{Figure 4}
\end{figure*}
\subsection{Model Architecture}
\Cref{Figure 4} illustrates the process of the suggested methodology. This demonstrates a comparative analysis between TL
models and ViT, with the application of Explainable AI (XAI)
techniques on the transfer learning models.
This study utilizes a range of pre-trained models grounded
in CNN architectures, such as VGG16, VGG19, ResNet50V2
and MobileNetV2 to enhance classification performance.
These models, initially trained on ImageNet, are subsequently
fine-tuned for the diagnosis of brain diseases utilizing MRI
data. Each model has unique benefits: VGG models employ 3x3 filters for detailed feature extraction, ResNet50V2
integrates residual connections to mitigate vanishing gradients, captures multi-scale features, improves efficiency through
depthwise separable convolutions, and MobileNetV2 is designed for optimal performance on resource-constrained devices [12]. The task is effectively modified for brain disease
illustrates the process of the suggested methodology.
This demonstrates a comparative analysis between TL models and ViT, with the application of Explainable AI (XAI) techniques on the transfer learning models.

This study utilizes a range of pre-trained models grounded in CNN architectures, such as VGG16, VGG19, ResNet50V2 and MobileNetV2 to enhance classification performance. These models, initially trained on ImageNet, are subsequently fine-tuned for the diagnosis of brain diseases utilizing MRI data. Each model has unique benefits: VGG models em ploy 3x3 filters for detailed feature extraction, ResNet50V2 integrates residual connections to mitigate vanishing gradi ents, captures multi-scale features, improves efficiency through depthwise separable convolutions, and MobileNetV2 is designed for optimal performance on resource-constrained devices\cite{salehi2023study}. The task is effectively modified for brain disease classification by fine-tuning pre-trained models, thereby retaining features acquired from large datasets. The architecture includes a flattening layer, dense layers with 1024 and 512 neurons, dropout layers to reduce overfitting and a softmax output for classification. Fine-tuning enhanced model efficacy and diminished training duration by utilizing pretrained ImageNet weights. This method has competently addressed the complexity of brain MRI categorization with enhanced efficiency.

\begin{figure}[ht]
\centering
\includegraphics[width=\linewidth]{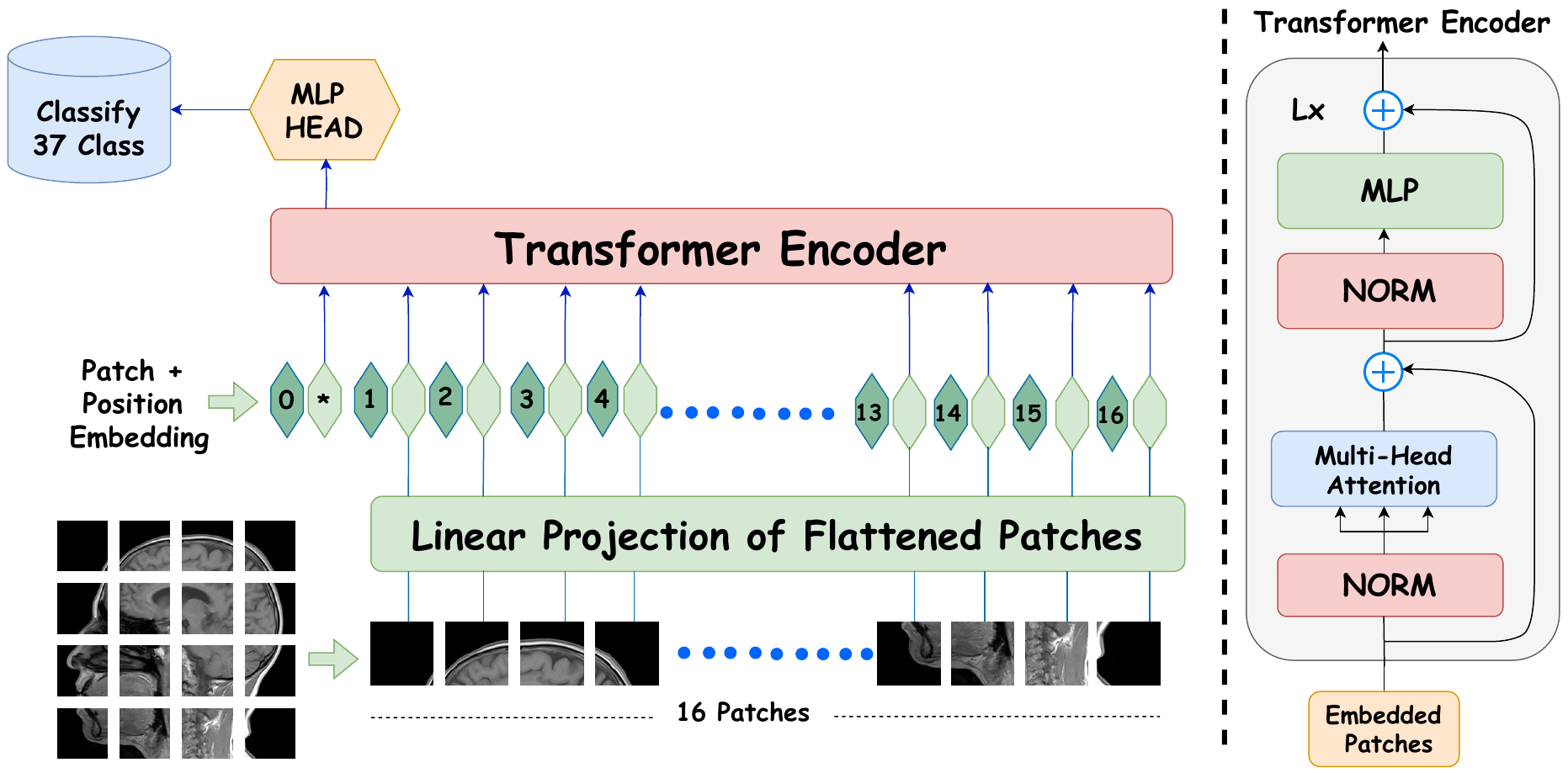}
\caption{Vision Transformer Architecture}
\label{Figure 5}
\end{figure}

The Vision Transformer (ViT) \cite{tummala2022classification}, analyzes images by segmenting them into smaller patches, similarly to how the original transformer processes words in natural language processing . In contrast to conventional transformer models that employ both an encoder and a decoder, ViT exclusively utilizes the encoder. An input image \(I\) is represented as \(\mathbb{R}^{H \times W \times C}\), where \(H\), \(W\), and \(C\) correspond to the height, width, and number of channels of the image, respectively. The image is divided into \(N\) patches, each of size \(P \times P \times C\), where:\(N = \frac{HW}{P^2}\). These patches are then flattened and embedded with positional information to preserve their spatial arrangement. Figure \ref{Figure 5} shows the architechture of the Vision Transformer(ViT).
A trainable classification token is incorporated into the patches, enabling final classification by a Multilayer Perceptron (MLP) head. The combined patches and positional embeddings are fed into the transformer encoder, which alternates between Multi-head Self-Attention (MSA) layers and MLP blocks. The MSA enables the model to discern relationships among patches, whereas the MLP has fully connected layers activated by the Gaussian Error Linear Unit (GELU).

The MSA outputs are derived by concatenating several attention heads, followed by processing through a final linear layer. The ViT's capacity to collect both local and global information in images renders it appropriate for applications such as medical imaging and object recognition, where comprehension of intricate details and long-range connections is essential. This method renders ViT a viable substitute for convolutional neural networks (CNNs) in diverse picture categorization tasks\cite{tummala2022classification}.

This study has employed a state-of-the-art ViT model with critical hyperparameters optimized for peak performance. The batch size of 256 is used here. Input images (128x128 pixels) are segmented into 64x64 patches and mapped into a 64-dimensional space. The model  comprises eight transformer layers, each with four attention heads, while the MLP head incorporates two dense layers with 2048 and 1024 units respectively. To prevent overfitting, we have implemented a dropout rate of 0.1 in the transformer layers and 0.5 prior to the final classification layer.

\section{Experimental Setup \& Result Analysis }
\label{Result Analysis}
\subsection{Experimental Setup}
The experimental configuration includes a machine featuring an Intel Core i5 13400F processor, an NVIDIA GeForce RTX 3060 12GB GPU, and 16GB of DDR4 RAM. All models are executed with Keras\cite{chollet2015keras}, a high-level API for neural networks, with TensorFlow \cite{tensorflow2015-whitepaper} serving as the backend. To guarantee robustness, each model is executed five times, with the optimal results chosen for analysis. For the Vision Transformer, the learning rate is set to \(1 \times 10^{-4}\), using the Adam optimizer, over 400 epochs, and the loss function applied was Sparse Categorical Crossentropy. In the TL models, a learning rate of \(1 \times 10^{-5}\) have been employed, with Adam as the optimizer and an identical loss function.

\subsection{Result Analysis}
The results from the study clearly demonstrate the superior performance of the Vision Transformer (ViT) model compared to traditional convolutional models like VGG16, VGG19, ResNet50V2 and MobileNetV2. Table \ref{tab:Table1} indicates that ViT has attained the maximum accuracy of 94.39\%, surpassing all other models, with ResNet50V2 following at 91.41\%. The transformer-based architecture of ViT enables the capturing of long-range dependencies in data, offering a considerable advantage in feature extraction and model generalization. This is especially evident compared to convolutional models that rely more on local feature detection. The superior accuracy of ViT highlights its ability to better understand complex patterns in the dataset, leading to more precise predictions.
\begin{table}[htbp]
\centering
\caption{Performance Metrics for Different Models}
\label{tab:Table1}
\renewcommand{\arraystretch}{1.7}
\begin{tabular}{|c|c|c|c|c|}
\hline
\textbf{Model} & \textbf{Accuracy} & \textbf{Precision} & \textbf{Recall} & \textbf{F1 Score} \\ \hline
VGG16          & 0.8844            & 0.9103             & 0.8854          & 0.8947            \\ \hline
VGG19          & 0.8905            & 0.8951             & 0.9108          & 0.8995            \\ \hline
Resnet50V2     & 0.9141           & 0.9276             & 0.9389          & 0.9211            \\ \hline
MobilenetV2    & 0.8856            & 0.8953             & 0.8938          & 0.8915            \\ \hline
\textbf{ViT}   & \textbf{0.9439}   & \textbf{0.9651}    & \textbf{0.9644} & \textbf{0.9638}   \\ \hline
\end{tabular}
\end{table}

\begin{table}[]
\centering
\caption{Comparison of Our Proposed Approach with Existing Methods}
\label{Table 2}
\renewcommand{\arraystretch}{1.1}
\begin{tabular}{|c|c|c|c|}
\hline
\textbf{Authors} & \textbf{Approach} & \textbf{Accuracy} & \textbf{Use of XAI} \\ \hline
\multirow{3}{*}{Brima et al.\cite{brima2021deep}} 
& \begin{tabular}[c]{@{}c@{}}Transfer Learning \\ of ResNet50\end{tabular} & 0.87 & \multirow{3}{*}{NO} \\ \cline{2-3}
& \begin{tabular}[c]{@{}c@{}}Transfer Learning \\ of ResNet50 \\ with Augmented Data\end{tabular} & 0.81 &  \\ \cline{2-3}
& \begin{tabular}[c]{@{}c@{}}Fine-tuning \\ ResNet50 with \\ Augmented Data\end{tabular} & 0.84 &  \\ \hline
Ahmed et al.\cite{majeed2024multi} 
& \begin{tabular}[c]{@{}c@{}}Transfer Learning \\ of MobileNetV3 with \\ Augmented Data\end{tabular} & 0.92 & NO \\ \hline

\multirow{2}{*}{\textbf{Our Proposed Method}} 
& \textbf{ViT} & \textbf{0.94} & \textbf{NO} \\ \cline{2-4}
& \textbf{ResNet50V2} & \textbf{0.91} & \begin{tabular}[c]{@{}c@{}}\textbf{YES}\\ \textbf{(Gradient-Based)}\end{tabular} \\ \hline
\end{tabular}%
\end{table}

\begin{figure}[h]
    \centering
    \begin{subfigure}[b]{0.46\textwidth}
        \centering
        \includegraphics[width=\linewidth]{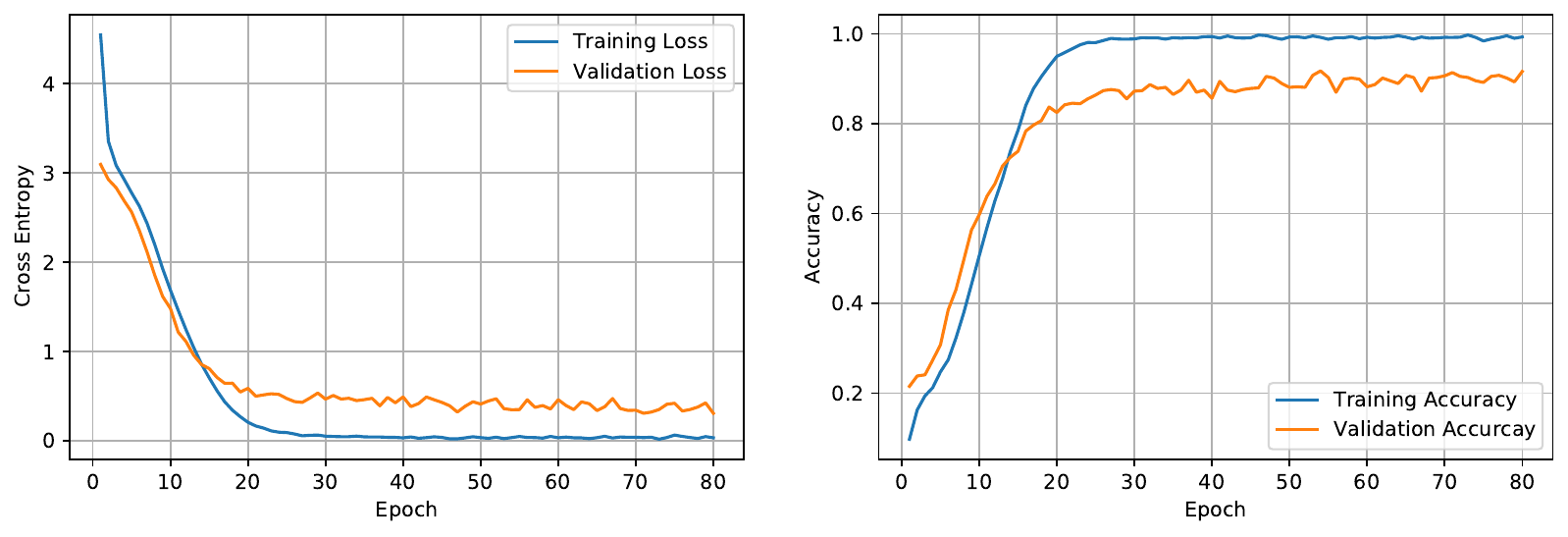} 
        \caption{ResNet50V2}
        \label{fig:subfig1}
    \end{subfigure}
    \hspace{0.02\textwidth} 
    \begin{subfigure}[b]{0.46\textwidth}
        \centering
        \includegraphics[width=\linewidth]{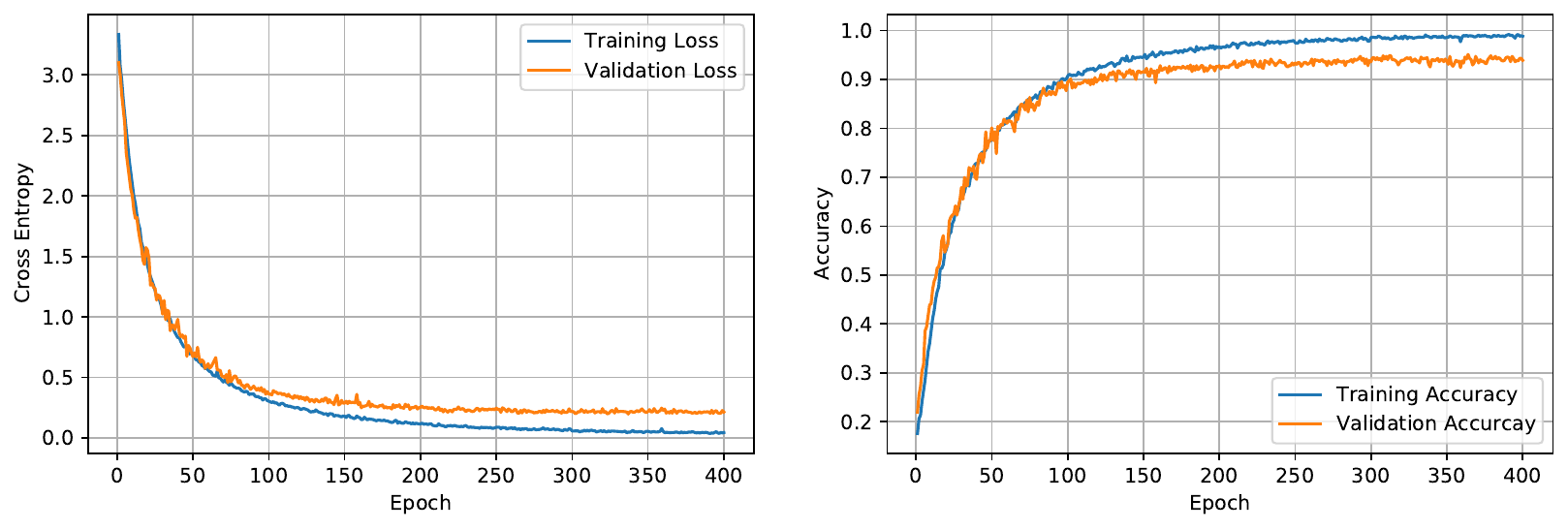} 
        \caption{Vision Transformer (ViT)}
        \label{fig:subfig2}
    \end{subfigure}
    \caption{Comparison of Training and Validation Loss Curves}
    \label{Figure 6}
\end{figure}

The training and validation loss curves, shown in Figure \ref{Figure 6}, further emphasize the efficiency of ViT. ViT not only converges faster during training but also maintains a lower validation loss compared to ResNet50V2. This indicates that ViT generalizes better to unseen data, with a reduced risk of overfitting. In contrast, the higher validation loss of ResNet50V2 suggests that it struggles more with generalization. Table \ref{Table 2} presents a comparative analysis of our proposed approach with existing methods in the field.

Overall, the results confirm that the Vision Transformer (ViT) significantly outperformed traditional CNN models like VGG16, VGG19, ResNet50V2 and MobilenetV2 achieving the highest accuracy of 94.39\%. ViT’s transformer-based architecture enables better generalization and feature extraction, especially in complex image classification tasks, where it excels in capturing long-range dependencies. In contrast, while CNN models show solid performance, they have struggled more with generalization and are less effective in capturing global features across the dataset.

\section{Gradient Based Explainable AI}
\label{Gradient Based Explainable AI}

XAI approaches aim to enhance the interpretability of AI models by revealing the mechanisms through which transfer learning models process and analyze data to generate predictions. \textbf{GradCAM}, a pivotal technology in XAI, offers visual elucidations for convolutional neural networks (CNNs) by employing target class gradients to pinpoint significant picture regions that affect predictions. It produces a heatmap to emphasize these regions, enhancing model transparency and interpretability without modifying the architecture, rendering it beneficial for tasks such as picture categorization, Visual Question Answering(VQA), and captioning\cite{rafi2022deep}. \textbf{GradCAM++} enhances this by overcoming constraints in identifying multiple instances of the same category. It integrates second-order derivatives, yielding enhanced object boundaries and augmenting heatmap precision. Conversely, \textbf{ScoreCAM} employs a gradient-free methodology, generating class-specific maps by assessing the impact of alterations in the activation maps on the resulting score. This technique improves visual clarity and is particularly helpful in pinpointing essential areas for picture categorization jobs. \textbf{FasterScore-CAM} enhances the procedure by concentrating on the most active channels within feature maps. This diminishes processing requirements while preserving interpretability, proving it beneficial for extensive datasets and intricate models.
\textbf{LayerCAM}, an advanced methodology, produces class activation maps using multiple layers of the CNN, rather than exclusively from the terminal convolutional layer. This enhances comprehension of spatial localization and object characteristics, consequently augmenting the accuracy of both object localization and classification\cite{faria2024explainable}.
\par Gradient-based XAI techniques encounter difficulties with  ViT due to architectural imbalances, including the absence of convolutional layers. ViT processes images as patch sequences and rely on self-attention mechanisms, which makes it more difficult to apply XAI methods on spatial feature maps and gradients. ViT analyzes images in segments, complicating the application of traditional CAM techniques. Furthermore, interpretability may be hampered by attention scores in ViT that do not correspond to human-intuitive relevance \cite{stassin2023explainability}. For this reason, instead of ViT five different XAI approaches are used in  ResNet50V2 model, which has outperformed all other TL models in evaluating brain MRI data. 

\begin{figure}[ht]
    \centering
    \includegraphics[width=\linewidth]{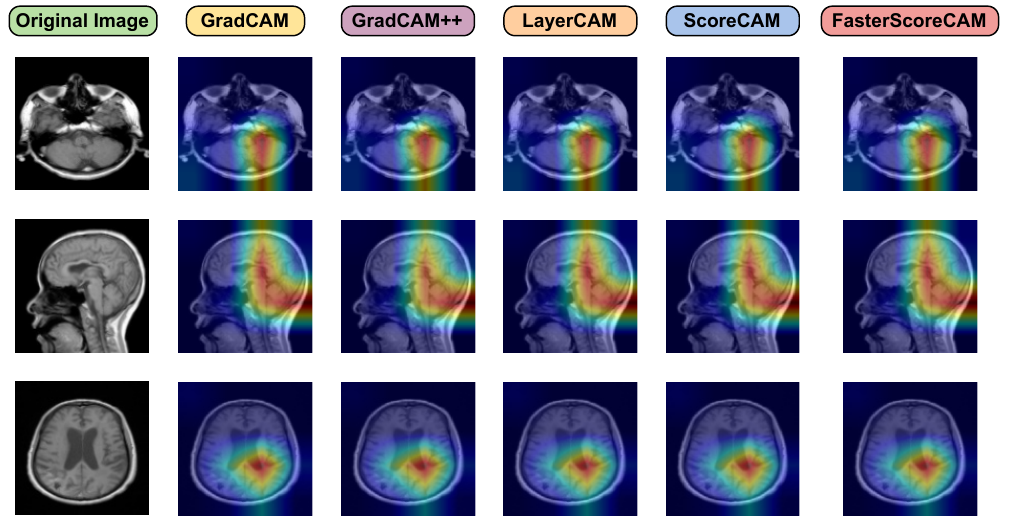}
    \caption{Gradinat Based XAI Visualization of Resnet50V2}
    \label{Figure 7}
\end{figure}

Figure \ref{Figure 7} highlights the consistent activation patterns seen across the ResNet50V2 model. This model demonstrates that GradCAM and GradCAM++ consistently highlight critical areas, whereas LayerCAM provides a broader yet still focused depiction.These methods employ feature maps, generally obtained from the intermediate or final layers preceding the flattening stage. GradCAM consistently emphasizes significant activations in essential brain regions, whereas GradCAM++ provides more concentrated and accurate heatmaps. LayerCAM records extensive activations while maintaining focus on critical regions. ScoreCAM allocates attention more uniformly throughout the brain image. In contrast, FasterScoreCAM concentrates on particular regions exhibiting elevated confidence levels. These methods consistently indicate significant regions, demonstrating that the models reliably recognize alike areas of interest across several XAI methodologies. XAI dramatically improves the reliability and interpretability of AI models through improved visual explanations, which is essential in vital domains like brain disease detection.

\section{Limitations and Future Work}
\label{Limitations and Future Work}
Despite the strength of the Vision Transformer (ViT) in classifying 37 distinct brain MRI classes, there are still areas for future work. One such area is use of Generative Adversarial Networks (GANs), which could be employed to generate synthetic MRI images and further test the model’s robustness, ensuring it generalizes well to more varied scenarios. Although the dataset is diverse and comprehensive, incorporating GANs could help justify the model’s true accuracy by challenging it with additional data variations. Practical implementation in clinical settings also presents challenges, such as computational resource demands and integration with existing hospital workflows. Future efforts will focus on incorporating GANs for data augmentation and validation, exploring hybrid architectures that blend the strengths of ViTs and CNNs, and optimizing the model for real-time deployment in healthcare environments.
\section{Conclusion}
\label{Conclusion}
In conclusion, this study demonstrates the significant potential of Vision Transformers (ViTs) for brain MRI classification, particularly in capturing global dependencies and improving overall accuracy in medical imaging tasks. By applying transfer learning and utilizing explainable AI (XAI) techniques, ViTs have been shown to outperform traditional CNN-based models like VGG16, VGG19, and ResNet50V2, achieving superior accuracy and generalization capabilities. The integration of XAI methodologies provided crucial insights into model predictions, making the classification process more transparent and interpretable. These findings underscore the importance of transformer-based architectures in medical image analysis and highlight the effectiveness of explainable AI in ensuring the reliability and trustworthiness of deep learning models in critical applications like brain disease diagnosis.
\bibliographystyle{unsrt} 
\bibliography{references.bib}

\end{document}